\documentclass[preprint,12pt]{article}

\usepackage{epsfig}

\usepackage{amssymb}

\usepackage{graphicx}
\usepackage{amsmath}
\usepackage{amsfonts}
\usepackage{amsbsy}
\usepackage{subfigure}
\usepackage{url}

\begin{document}


\title{Enforcing Label and Intensity Consistency for IR Target
Detection\footnote{First appeared in OTCBVS 2011~\cite{parag11otcbvs}. This manuscript presents updated results and an extension.}}


\author{Toufiq Parag\\ {\small Janelia Farm Research Campus-HHMI, Ashburn, VA 20147}}

\date{}
\maketitle

\begin{abstract}
This study formulates the IR target detection as a binary
classification problem of each pixel. Each pixel is associated with
a label which indicates whether it is a target or background pixel.
The optimal label set for all the pixels of an image maximizes
aposteriori distribution of label configuration given the pixel
intensities. The posterior probability is factored into (or
proportional to) a conditional likelihood of the intensity values
and a prior probability of label configuration. Each of these two
probabilities are computed assuming a Markov Random Field (MRF) on
both pixel intensities and their labels. In particular, this study
enforces neighborhood dependency on both intensity values, by a
Simultaneous Auto Regressive (SAR) model, and on labels, by an
Auto-Logistic model. The parameters of these MRF models are learned
from labeled examples. During testing, an MRF inference technique,
namely Iterated Conditional Mode (ICM), produces the optimal label
for each pixel. The detection performance is further improved by
incorporating temporal information through background subtraction.
High performances on benchmark datasets demonstrate effectiveness of
this method for IR target detection.

\end{abstract}

%
%

\section{Introduction}
\begin{figure}
\begin{center}
\subfigure[]{\includegraphics[width=1.35in,height=1.15in]{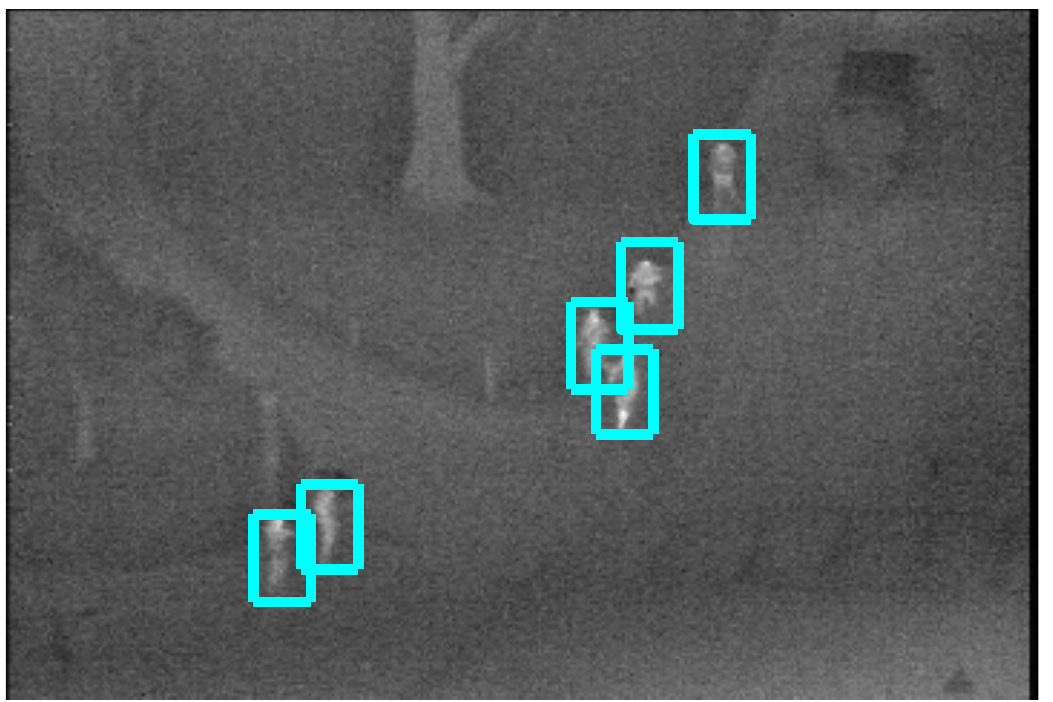}}
\subfigure[]{\includegraphics[width=1.35in,height=1.15in]{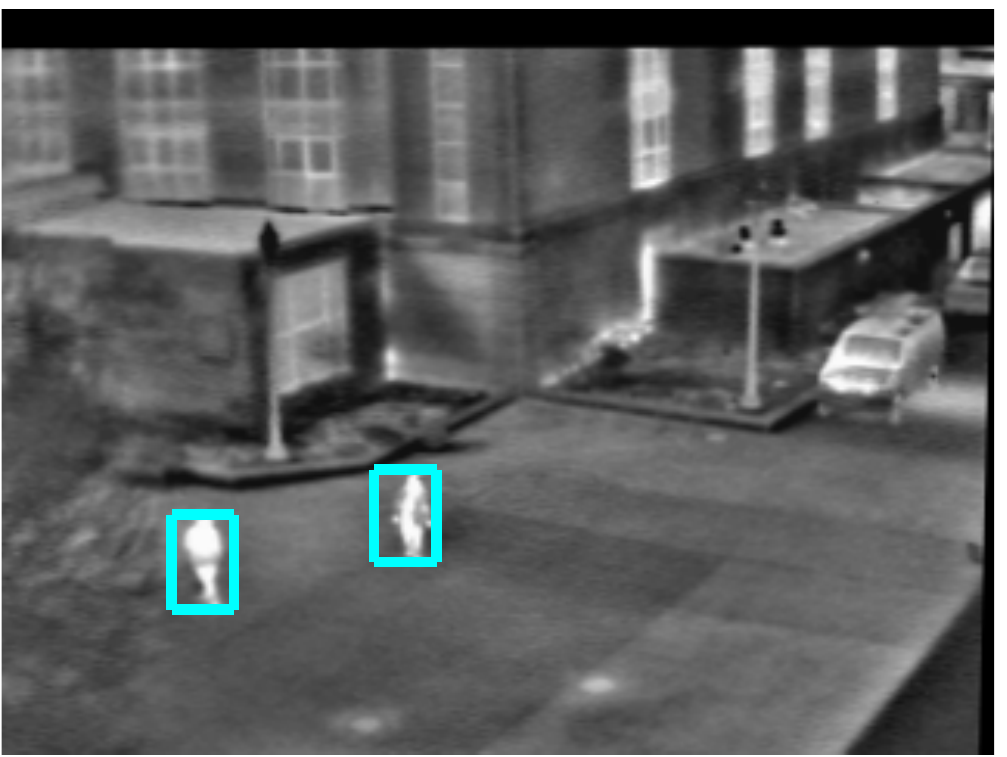}}
\subfigure[]{\includegraphics[width=1.35in,height=1.15in]{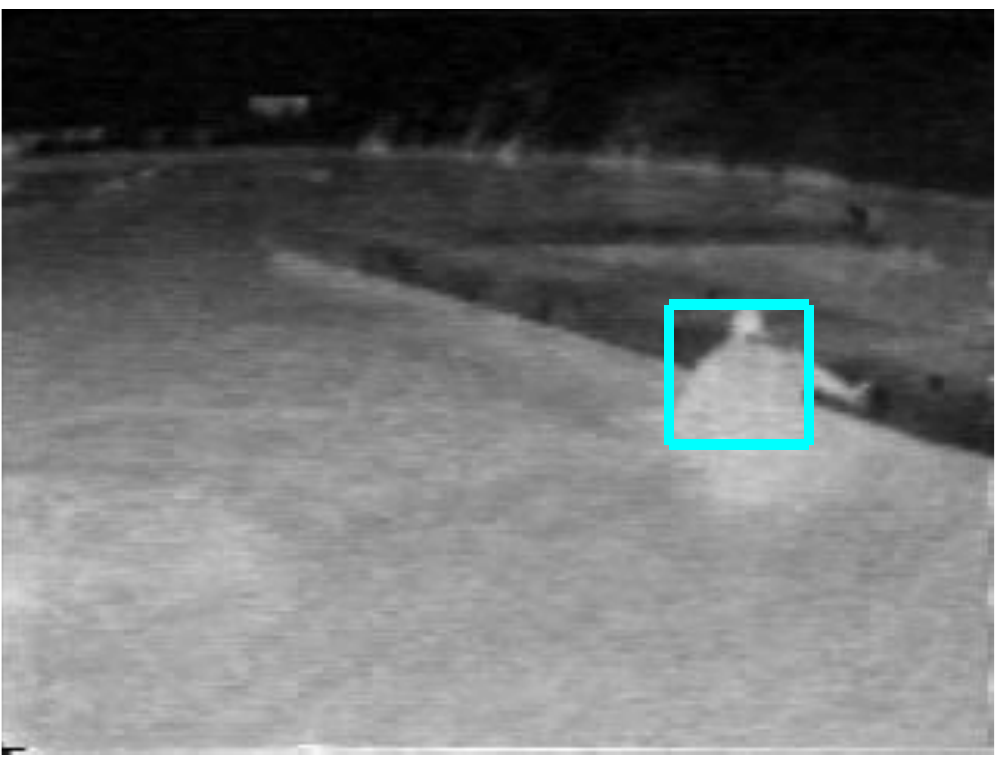}}\\
\subfigure[]{\includegraphics[width=1.35in,height=1.15in]{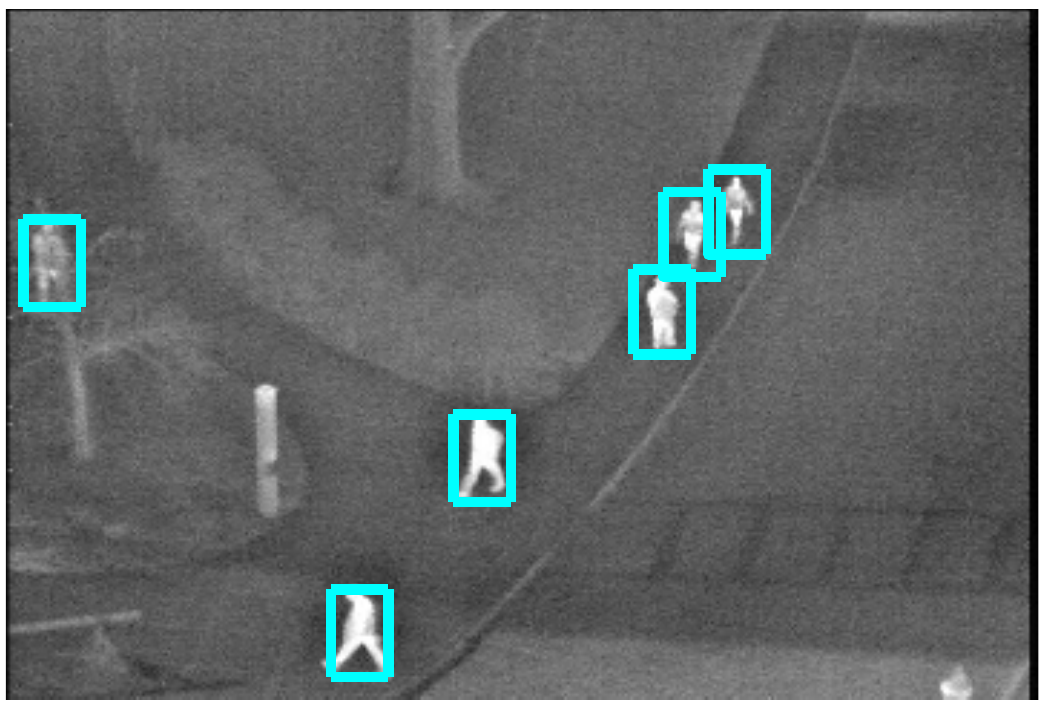}}
\subfigure[]{\includegraphics[width=1.35in,height=1.15in]{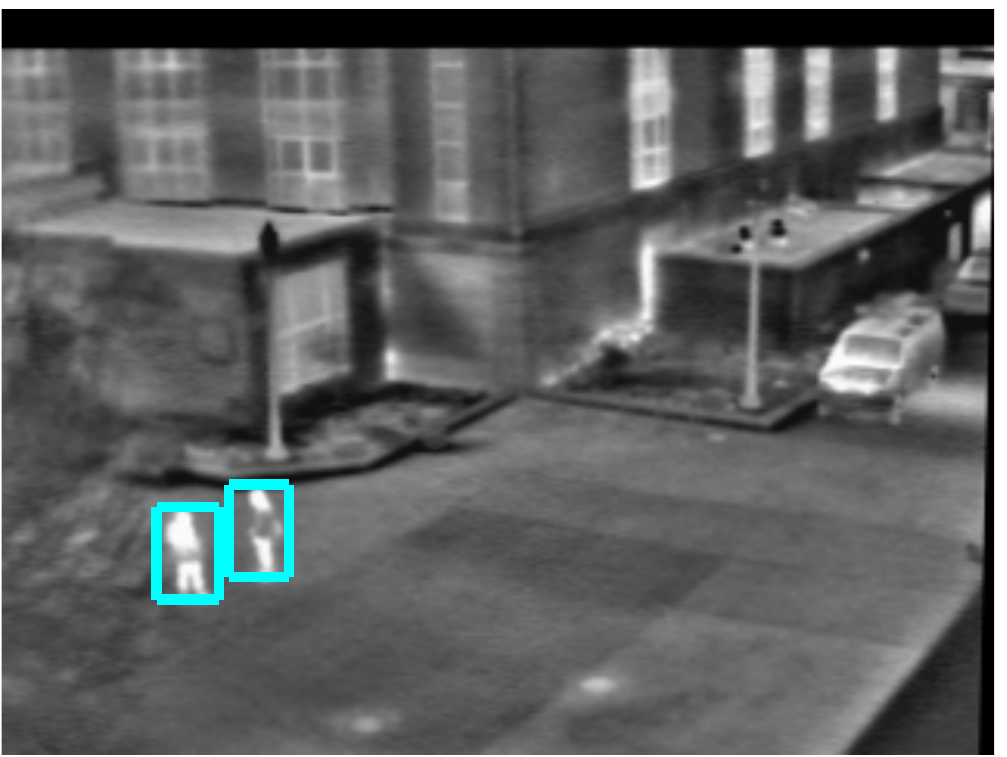}}
\subfigure[]{\includegraphics[width=1.35in,height=1.15in]{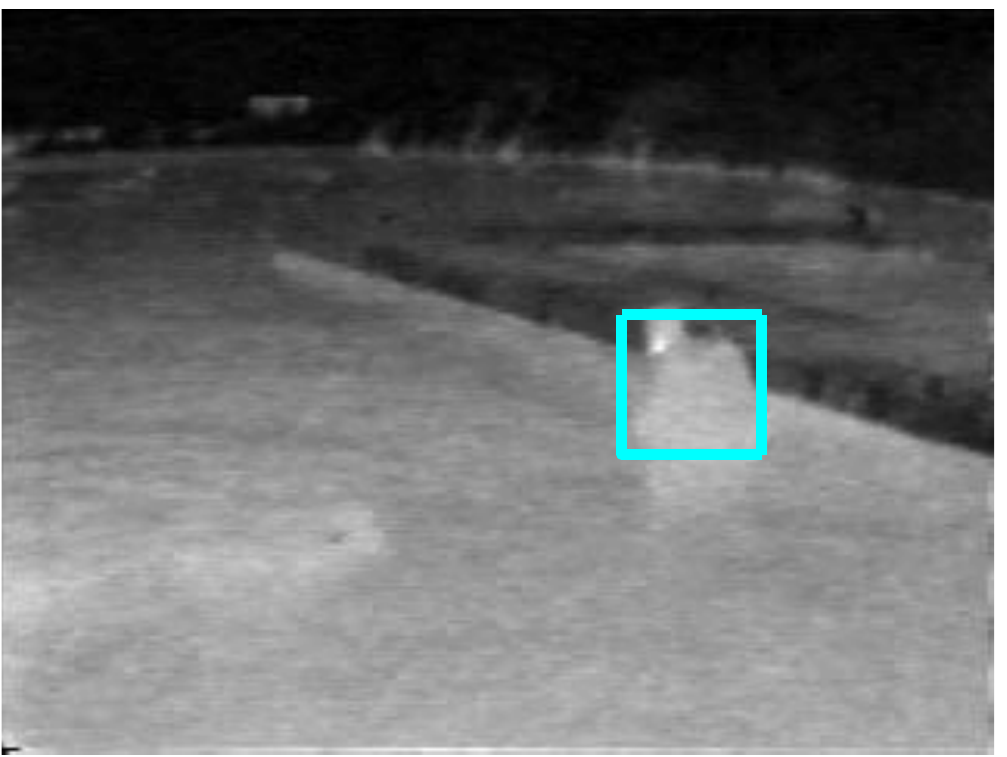}}
\end{center}
\caption{\footnotesize Sample input images for IR target
detection.}
\label{F:INTRO}
\end{figure}

Infrared (IR) images provide valuable information for visual
surveillance. Since IR wavelength responds to heat, it can indicate
the locations of objects in an image taken during the day or at
night. Hence, these images can be exploited in surveillance system
that need to be executed around the clock or in scenarios where
visual features are not discriminative enough for detection. But,
detecting objects in IR images is not trivial. Targets are often not
as illuminated as one would expect in ideal case. To deteriorate the
situation, luminous background regions also appear frequently in
thermal images. We show several sample images in
Figure~\ref{F:INTRO} where (groundtruth) target locations are
displayed in rectangles.

The images (a) and (d) in Figure~\ref{F:INTRO} indicate that the
distribution pixel intensities within target region is multimodal.
Comparing these images with (b) and (e), one could realize that the
distribution also overlaps with that of background pixel values. The
overlap is even more evident in images (c) and (f) that show a
swimmer/diver swimming coming out of the water. These images
strongly suggest that the same object detector can not work
satisfactorily for all these scenarios -- we need to learn the
object pattern for each of these cases. Furthermore, it is also
evident that independently pixel intensities are not discriminative
features for target classification. These characteristics are also
observed by past studies which uses different other low level
features as described in the next paragraph. It is worth noting
that, though the individual pixel values are not very
discriminative, intensity arrangement within an image patch seems to
be highly informative and could potentially result in superior
performance if modeled appropriately.

Detecting pedestrians in natural images (i.e., visible spectrum) has
gained much attention in past several
years~\cite{leibe08}\cite{viola03iccv}\cite{dalal05cvpr}. Several
previous works on IR object detection extended algorithms for
pedestrian detection in visible spectrum to thermal images. The
study by Zhang et. al.~\cite{zhang07otcbvs} experimented with the
Edgelet and Histogram of Oriented Gradient (HOG) and applied
Adaboost and Support Vector Machine (SVM) classifiers to identify
pedestrian in IR images. The work in~\cite{jungling09otcbvs} uses
the SURF features and utilizes a voting scheme of detected codewords
to determine object location in image. A more recent
work~\cite{venkataraman10otcbvs} computes composite combinatorial
features from low level features such as intensities and HOG to
represent object pattern and  employs a boosting classifier for
detection. Though these studies indirectly learn the pattern of
object patch (as described by the features they use), they do not
explicitly model the relationship of the pixel values and their
labels with neighboring ones. More specifically, the labels of the
pixels are considered to be independent of each other.

Other approach such as~\cite{davis04otcbvs} proposes a method to
refine the contours of foreground regions, detected by a background
subtraction model, to determine the target locations. Temporal
information between consecutive frames, is utilized in IR object
detection as well. Wang et. al.~\cite{wang06otcbvs} combines the
results of detection and tracking to extract target locations from
scene. Temporal information is shown to assist identifying small
objects and to remove many false positive locations otherwise
detected in single frame methods. In~\cite{leykin06otcbvs}, the
authors combines thermal and visible information and adopt a
Particle Filtering approach for tracking (multiple) people in the
image. An earlier work~\cite{dai05otcbvs} of similar spirit adds
shape information with appearance to identify objects foreground
regions.

This paper proposes to model both the pixel intensities and labels
to be dependent upon those of its neighbors. The optimal label
configuration of image pixels is supposed to maximize the posterior
probability given the intensity configuration. This posterior can be
factored to a conditional likelihood of intensity configuration and
a prior probability of the label configuration. We model both the
conditional likelihood of intensity arrangement (given the label
arrangement) and prior probability of labels using (two different
types of) Markov Random Field (MRF)~\cite{li01book}. As a result,
both the label and intensity of a pixel becomes related to those of
its neighbors. This relationship is learned from given examples of
target and background patches. Intuitively, we are learning the
intensity and label correlation patterns in target regions of IR
images. An MRF inference algorithm, namely the Iterated Conditional
Modes (ICM), determine the optimal value of pixel labels during test
phase.

In a similar spirit of~\cite{davis04otcbvs, wang06otcbvs}, we also
combine temporal information in order to refine the detection
performance. The result of a background subtraction algorithm is
fused with that of the proposed method to generate the final
detection output. Background subtraction and change detection
techniques have a relatively longer history than other low level
vision problems in computer vision literature, see~\cite{parag06}
and the references therein. Most of these studies represent the
temporal history of background pixel intensities by a parametric or
non-parametric statistical model; pixel locations whose intensity
significantly deviates from this model are classified as foreground
pixels. Although the background subtraction method alone can not
produce a satisfactory performance~\cite{davis04otcbvs}, we show
that they can assist the proposed method to achieve the maximal hit
rate with zero false detections.

Since its introduction, MRF models have been applied to many
problems in vision. Due to recent discoveries of elegant and
efficient inference algorithms, MRF models have become widely
popular in vision. Some notable applications are image
segmentation~\cite{boykov01jolly}\cite{rother04grabcut}, object
recognition~\cite{kumar05objcut}, stereo matching, feature
correspondence~\cite{torresani08}, image denoising and
reconstruction~\cite{roth-05foe}, clustering~\cite{zabih04} etc.
Surprisingly enough, MRF has not yet been used for IR target
detection to the best of our knowledge.

Coupled MRF models were also used in~\cite{derin86grf}
and~\cite{simchony88} for texture segmentation of images. We show
the applicability of similar framework on IR images for target
detection. The specific instances of MRF models assumed in these
studies are different from those incorporated in this paper. There
exist more sophisticated frameworks that represent the pixel
intensity arrangement within a patch by MRF models,
e.g.,~\cite{manjunath90} or more recent Fields of Expert
model~\cite{roth-05foe}.

The paper is organized as follows: Section~\ref{S:OVERALL} describes
the overall probability structure followed by our method.
Section~\ref{S:MRFS} describes the specific MRFs imposed on
intensity and label layers. Techniques for learning the parameter
values and for computing the optimal labels of test images are
discussed in Sections~\ref{S:TRAIN} and~\ref{S:TEST} respectively.
We report the results of our method on benchmark datasets in
Section~\ref{S:RESULTS} before concluding our findings in
Section~\ref{S:CONC}.

\section{MAP Estimation of Target Locations}\label{S:OVERALL}

\begin{figure}[h]
\begin{center}
\subfigure{\includegraphics[width=1.5in,height=1.25in]{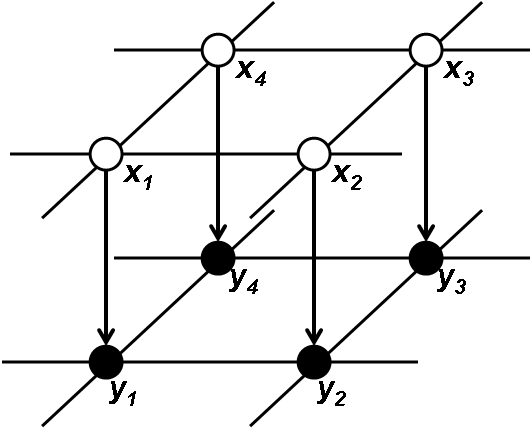}}
\end{center}
\caption{\footnotesize Bi-layer MRF model.} \label{F:BN}
\end{figure}

Let $\mathbf{y} = \{ y_i \in \mathbb{R} ~|~ i = 1, \dots, n\}$ be
the intensities of an IR image. We wish to label each location $i$
in the image to be a background or target pixel. Let $\mathbf{x} =
\{ x_i \in \{0, 1\} ~|~ i = 1, \dots, n\}$ be the class labels for
pixels in this image : $x_i = 1$ and $0$ implies a target and
background pixel respectively. Then, our goal is to compute the
optimal label configuration $\mathbf{x}^*$ that maximizes the
aposteriori distribution $p(\mathbf{x} ~|~ \mathbf{y}) ~\propto~
p(\mathbf{y} ~|~ \mathbf{x} )~p(\mathbf{x}).$ In the proposed
framework, both the conditional likelihood $p(\mathbf{y} ~|~
\mathbf{x})$ and the prior $p(\mathbf{x})$ are modeled by Markov
Random Field (MRF)s. Our objective is to compute the optimal values
of both $\mathbf{y}$ and $\mathbf{x}$ provided the intra-layer
relationship (captured by respective MRFs) and the inter-layer
dependence utilizing appropriate statical inference techniques. The
probabilistic dependencies among the intensities, class labels and
in between are depicted in Figure~\ref{F:BN}.

\section{MRF Models for Target Detection}\label{S:MRFS}
Markov Random Field (MRF) is a probabilistic graphical structure
designed to model the dependence among (neighboring) datapoints. In
MRF, a random variable is considered to be probabilistically
dependent only to its neighbors. There are several different
definitions of neighborhood systems, for example, a pixel in an
image is a neighbor of all other pixels it is connected to in
4-connectivity. A pairwise (or higher order) relation among the
neighbors is defined as clique within a neighborhood. Given a
suitable definition of neighborhood and clique, the joint
probability of a set of variables $\boldsymbol{\omega}$ is given by
\begin{equation}\label{E:MRFDEFN}
p(\boldsymbol{\omega}) = {1 \over Z}~\exp \Bigl \{-\sum_{c \in
{\cal C}} E_c(\boldsymbol{\omega}_c)\Bigr \},
\end{equation}

where $Z$ is a normalizing term~\cite{li01book}. The set of
variables may either correspond to the class labels $\mathbf{x}$ or
to the intensity levels $\mathbf{y}$ (conditional on $\mathbf{x}$).
For each clique $c$ in the set $\cal C$ of all possible cliques, a
cost (or potential) function $E_c$ penalizes the disagreement among
the values of $\boldsymbol{\omega}_c$ associated with $c$. The
optimal $\boldsymbol{\omega}^*$ is the one that maximizes the joint
probability, or, in other words, minimizes the so called Gibbs
energy function $E(\boldsymbol{\omega}) = \sum_{c \in {\cal C}}
E_c(\boldsymbol{\omega}_c)$. Definition of neighborhood and clique
varies with application, e.g., considering the pixel arrangement in
an image a a grid,  a 4-connectivity defines a first order
neighborhood and all possible pairs containing the reference
variable are defined as cliques~\cite{li01book}.

We use a functionally homogenous potential function, that is $E_c =
E$ for all $c \in {\cal C}$. Following two sections discuss the MRF
models for the conditional likelihood and prior distribution
respectively.

\subsection{MRF for Conditional Likelihood}

We model the conditional probability $p(\mathbf{y}| \mathbf{x})$ of
intensities at all pixel locations by an MRF. That is, our model
imposes a dependence among intensity $y_i$ at each pixel to those
$y_j,~ j\in {\cal N}_i$, in the neighborhood ${\cal N}_i$ of pixel
$i$. The conditional likelihood $p(\mathbf{y}~|~\mathbf{x})$ is
factored into locally dependent probabilities by assuming a
Simultaneous Auto Regression (SAR) MRF over the intensities.
Essentially, we are assuming that the pixel intensity of any
specific location is dependent to those of its neighboring
locations. In SAR model, the deviation between intensity at location
$i$ and its mean is expressed by linear combination of the
deviations of its neighboring pixels perturbed by Gaussian noise. It
should be clarified here that, the pattern of interaction among the
neighboring pixels will be different for the target and background
class. Therefore, in order to represent the neighborhood
relationship, we need two SAR MRFs: one for the target and the other
for the background class.

In our framework, we assume all the pixels of a particular class $l
\in \{\text{target, background}\}$ are generated from same process
with mean $\mu^l$. The noise variance is also assumed to the same
$(\sigma^{l})^2$ for each class. The pixel value $y_i$ at location
$i$ is related to its neighbors in SAR model for class $l \in
\{\text{object, background}\}$ by the following equation.

\begin{equation}\label{E:SAR_YI}
y_i = \mu^l + \sum_{j\in{\cal N}_i} \beta^l_{ij} (y_j - \mu^l) +
\epsilon.
\end{equation}

The parameters $\beta^l_{ij}$ determines the influence of the
neighboring pixels $j \in {\cal N}_i$ on pixel $i$. The value of
$\beta^l_{ij}$ will be different for different $j$, i.e., different
neighbor locations within the neighborhood. The noise is assumed to
be Gaussian with zero mean and standard deviation $\sigma^l.$ The
conditional probability of $y_i$ given only its neighbors, which
will be useful in computing MRF inference, is therefore follows a
Normal distribution with mean $\mu^l + \sum_{j\in{\cal N}_i}
\beta^l_{ij} (y_j - \mu^l)$ and variance $(\sigma^{l})^2$

\begin{equation}\label{E:SAR_PROB} p(y_i ~|~ x_i = l,
y_{{\cal N}_i}) = {1 \over (2 \pi)^{1/2} \sigma^l}~
\exp  \Bigl [ - {1\over 2(\sigma^{l})^2} \bigl \{ y_i - \mu^l
- \sum_{j\in{\cal N}_i} \beta^l_{ij} (y_j - \mu^l)\bigr \}^2
\Bigr ] .
\end{equation}

In the following two equations, we state the pixelwise and pairwise
(clique sizes 1 and 2) potential functions of this SAR model to
interpret how they act on pixel values within a neighborhood.

\begin{align}\label{E:SAR_E}
    E^s(y_i) &= { (y_i - \mu^l)^2 \over 2(\sigma^{l})^2}, \\
E^s(y_i, y_j) &= (\beta_{ij}^{l})^2 ~{(y_i - \mu^l)(y_j - \mu^l)
\over 2 (\sigma^{l})^2}.
\end{align}

The pointwise potential function $E^s(y_i)$ penalizes large
difference between pixel values and their mean. Weighted by
$\beta_{ij}^l$ values,  the pairwise potential $E^s(y_i, y^j)$
penalizes the discrepancy in intensity deviation from corresponding
mean in neighboring pixels. The pairwise energy will be minimum when
both the neighboring pixels have low deviation.

More sophisticated models of~\cite{roth-05foe}~\cite{manjunath90}
are rich and therefore are expected to describe the pattern more
accurately. However, they also require complicated learning and
inference algorithms that are approximate and computationally
expensive. As we will see in the results section, the simpler SAR
model described in this section is capable of producing high
detection ratios on different IR experiments.

\subsection{MRF for Prior}
The prior distribution $p(\mathbf{x})$ of labels is modeled by an
Auto-Logistic MRF. The potential functions for Auto-Logistic model
are as follows.

\begin{align}\label{E:AUTO_E}
    E^a(x_i) &=  \nu_i x_i, \\
    E^a(x_i, x_j) &=  \gamma_{ij}~x_i x_j.
\end{align}

\begin{figure}[h]
\begin{center}
\subfigure[Lowest
cost]{\includegraphics[width=1.2in,height=1.25in]{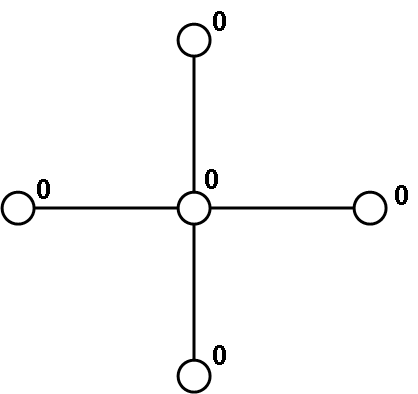}
}\qquad
\subfigure[Moderate
cost]{\includegraphics[width=1.2in,height=1.25in]{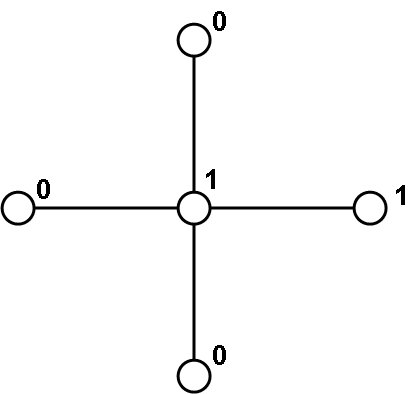}
}\qquad
\subfigure[Highest
cost]{\includegraphics[width=1.2in,height=1.25in]{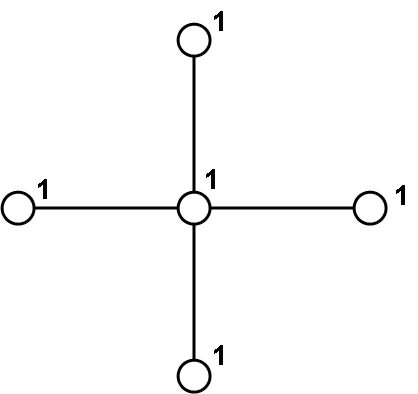}}
\end{center}
\caption{\footnotesize Example labeling of a pixel and its
neighbors in a 4-way connectivity. The lowest cost is incurred
when the pixel and all its neighbors are labels as 0
(background), as shown in leftmost figure. Auto model cost
increases as more pixels are labeled as targets. }
\label{F:AUTO_ILLUS}
\end{figure}

Clearly, the pointwise potential function $E^a(x_i)$ grows with the
number of pixels labeled as target class. The pairwise potential
function $E^a(x_i, x_j)$ accumulates a certain value $\gamma_{ij}$
whenever the neighboring pixel labels $x_i$ and $x_j$ both belong to
the target class (recall $x_i \in \{0, 1\}$).  This is illustrated
in Figure~\ref{F:AUTO_ILLUS} which demonstrates how changing the
pixel label to 1 (target) increases clique cost.  Auto-logistic MRF
model tends to generate less detection while maintaining the label
similarity in nearby pixels. This model is particularly suitable for
target detection because typically targets are sparse and consume a
small area in the image.

The conditional probability of any label $x_i$ is given by

\begin{align}\label{E:AUTO_PROB} p(x_i ~|~ x_{{\cal N}_i}) =
&{\exp \bigl [ \nu_i x_i +
\sum_{j \in {\cal N}_i} \gamma_{ij} x_i x_j \bigr ] \over 1+
\exp \bigl [ \nu_i +
\sum_{j \in {\cal N}_i} \gamma_{ij} x_j \bigr ]} .
\end{align}

\section{Parameter Estimation}\label{S:TRAIN}
The parameters of SAR and Auto-Logistic models are learned from
labeled examples. The intensity mean for target and background
pixels are estimated from labeled pixels for both classes. Since SAR
model is defined by a set of equations in the form of
Equation~\ref{E:SAR_YI}~\cite{li01book}, the linear
coefficients$\beta_{ij}$ and noise variances $\sigma^{l~2}$ are
learned by Least Squares Linear Regression.

The parameters of the Auto-Logistic model was computed by the
Pseudo-Likelihood (PLL) method. The product of all locally
conditional probabilities is defined as the Pseudo-Likelihood in MRF
literature.

\begin{equation}\label{E:PLL}
pll(x) = \prod_{i} p(x_i ~|~ x_{{\cal N}_i}).
\end{equation}

PLL is considered to be an approximation of the likelihood values
for estimation purposes~\cite{li01book}. We applied a non-linear
programming algorithm to maximize the log PLL of Auto-Logistic
model, as stated below, in terms of the parameters.

\begin{equation}\label{E:LPLL}
lpll(x) = \sum_i \Bigl \{ \nu_i x_i + \sum_{j \ \in {\cal N}_i}
\gamma_{ij} x_i x_j \Bigr \} - \sum_i \log \Bigl \{ 1 + \exp
\bigl ( \nu_i + \sum_j \gamma_{ij} x_j \bigr ) \Bigr \}.
\end{equation}

\noindent In all our experiments, the nonlinear optimization
technique steadily converged to a solution in $15 \sim 20$
iterations.

\section{Inference}\label{S:TEST}
The optimal label configuration $\mathbf{x}^*$ is computed by
Iterated Conditional Mode (ICM) technique~\cite{besag86dirty} for
the proposed algorithm. The ICM algorithm works locally by
determining the optimal label at each location given the
observations and labels of its neighbors. In our scenario, the
conditional probability to maximize at each location is proportional
to the following:

\begin{equation}\label{E:ICM}
p(y_i ~|~ x_i , y_{{\cal N}_i})~p(x_i ~|~ x_{{\cal N}_i}).
\end{equation}
This is a product of conditional probabilities of pixel value and
its label given those of its neighbors that are modeled by SAR and
Auto-Logistic models (Equations~\ref{E:SAR_PROB}
and~\ref{E:AUTO_PROB}) respectively. The labels are updated by
$x^*_i = \arg\max_{x} p(y_i ~|~ x_i , y_{{\cal N}_i})~p(x_i ~|~
x_{{\cal N}_i})$ iteratively until convergence. Several update
schemes have been suggested for ICM in the literature. In our
experiments, we updated the pixel locations and its neighbors (4-way
connection) alternatively. This can be achieved by updating the
alternate pixel in both horizontal and vertical directions for a
4-connectivity neighborhood~\cite{besag86dirty}.

It is important to clarify here that, the purpose of inference in
our algorithm is to find the optimal label configuration
$\mathbf{x}^*$ and \emph{not} to reconstruct or denoise the image by
producing the optimal intensities at each location. The SAR MRF
model is utilized solely to calculate the conditional probabilities.

\section{Incorporating Temporal Information}

In a video or temporally contiguous IR images sequence, the pattern
of background intensity values of each image location can be
statistically modeled by background subtraction techniques. In an
attempt to reduce the amount of false positives, we combine the
output of the coupled MRF (SAR-Auto) framework with that of a
background subtraction method. To this end, we adopt the
nonparametric statistical model proposed in~\cite{elgammal00}. This
nonparametric method models the distribution of each pixel value
over a certain period of time through Kernel Density Estimation
(KDE) and computes a likelihood value for each pixel to belong to
the background class . This probability is then checked against a
threshold to generate the 0-1 decisions. Formally, let $x_i^t$ be
the intensity value at location $i$ at time $t$. The probability of
pixel $i$ to be a background location given $T$ previous
observations $x_i^t,~ t = 1, \dots, T$, is computed by KDE by the
following formula.

\begin{equation}\label{E:KDE}
p(i \in \text{bckgnd}) \propto {1 \over T} \sum_{t=1}^T \exp \Bigl [ {-1 \over 2 \sigma^2} (x_i - x_i^t)^2 \Bigr ]
\end{equation}

In this equation, the value of length of history $T$ and kernel
bandwidth $\sigma$ are two external parameters. The value of $p(i
\in \text{bckgnd})$ is discretized by a low threshold in order to
minimize the false negative rate. The output of the coupled MRF
model (or its variants, as explained later in experiments section)
and the background subtraction are combined by a logical AND
operations. Although more involved fusion strategy could be
performed in this step (and we encourage the reader to do so
whenever necessary), we found in our experiments that a logical AND
is sufficient to produce a high detection rate with very few or no
false positives.

\section{Experimental Results}\label{S:RESULTS}
We tested our method on three datasets of OTCBVS benchmark
data~\cite{otcbvs} : Dataset 01 and 03 with IR images of people
walking on the street and Dataset 05 with two divers coming out of
the water (referred to as IRD01, IRD03 and IRD05 resp.). IRD01
contains 261 IR images in 9 subsets (excluding Subset 3 which
contains inverse IR images) with relatively darker (colder)
background regions, see Figure~\ref{F:INTRO} (a) and (d)for sample
images. However, the brightness of targets also vary significantly
among different sequences, notably in Subsets 7, 9, 10. On the other
hand, the images in IRD03 and IRD04 (image sets 5a and 4a in Dataset
03 of OTCBVS) consists of images with brighter target and background
regions as shown in Figure~\ref{F:INTRO} (b) and (e).

IRD05 is particularly interesting for detection IR images since the
targets in this dataset are so called \lq cold\rq~ targets. That is,
the temperature of the two divers in these images are too low for
the IR camera to mark them with bright pixels
(Figure~\ref{F:INTRO}(c), (f)). As a result, pixels on the target
appear to be very similar to those in the background.

\begin{figure}[h]
\begin{center}
\subfigure[Sample training
image]{\includegraphics[width=1.5in,height=1.25in]{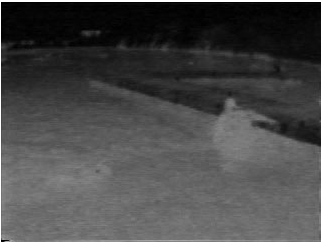}
}\qquad
\subfigure[Sample test
image]{\includegraphics[width=1.5in,height=1.25in]{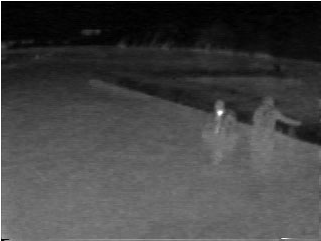}}
\end{center}
\caption{\footnotesize Sample training and test image for
IRD05.} \label{F:TRN1}
\end{figure}

For IRD01, IRD03 and IRD04, the parameters were learned from 50, 100
and 100 randomly selected images respectively. Due to scarcity of
targets in IR03, more images were necessary to gather enough
observation for learning. In IRD05 images, we learn the models from
50 images where only one target is visible and test the method on 70
images where both the targets are visible. Sample training and test
images of IRD05 are shown in Figure~\ref{F:TRN1}.

\begin{figure}[h]
\begin{center}
\subfigure[
]{\includegraphics[width=1.5in,height=1.25in]{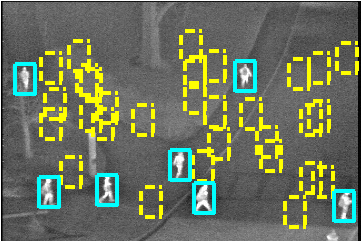}
} \qquad\qquad
\subfigure[ ]{\includegraphics[width=1in,height=1in]{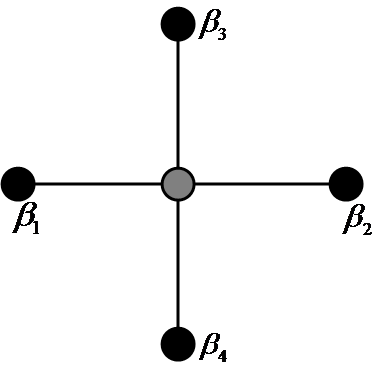}
}
\end{center}
\caption{\footnotesize (a) Bounding boxes around target and
background regions. (b) Four way connectivity.} \label{F:BB}
\end{figure}

The pixel intensities within the target bounding boxes are utilized
to estimate the parameters of target SAR model. For background SAR
model, we randomly selected background patches of same size as
targets and used pixels within for learning. Figure~\ref{F:BB}(a)
shows an image with target (solid) and background (yellow dashed)
bounding boxes overlaid on it. The neighborhood used for both SAR
and Auto-Logistic model is the 4-way associativity as shown in
Figure~\ref{F:BB}(b). In intensity SAR model, each of the four
connections are regarded as cliques and will be associated with
different mixing weights $\beta_{ij}^l$. The Auto-Logistic model on
labels was simplified by constraining the $\gamma_{ij}$ values for
four neighbors to be the same.

The built in generic solvers for linear regression and nonlinear
optimization problems on Matlab were used for parameter estimation
of SAR and Auto-logistic models respectively. Learning the models
required less than 5 minutes for all the datasets on an Intel
Pentium 1.83GHz PC with 1 GB RAM. During inference, a mex-C++
implementation of ICM converged to optimal labeling within 15
iterations and required approx 1 second per frame on average.

To retrieve the target location from the output of ICM, we utilized
the ratio of conditional probability values $\rho_i = {p(y_i ~|~ 1 ,
y_{{\cal N}_i})~p(1 ~|~ x_{{\cal N}_i}) \over p(y_i ~|~ 0 , y_{{\cal
N}_i})~p(0 ~|~ x_{{\cal N}_i})}$ each pixel $i$. These values are
thresholded at several different level $\delta_k,~k=1, \dots, K$.
Pixels with $\rho_i > \delta_k$ are congregated by a connected
component search and the center of each connected component is
considered to be the candidate locations. The number of detections
is further reduced by merging overlapping boxes. Bounding boxes with
$50\%$ overlap are merged together and the mean of their centers
become the center of the new detected location. Every detected
location having a $30\%$ overlap with a groundtruth target bounding
box is counted as a correct detection for performance evaluation.

In order to compare the contribution of each of the intensity (SAR)
and label (Auto-Logistic) MRF models, we experimented with two other
simplified methods on these datasets as follows.
\begin{enumerate}
  \item SAR-i model: represents the conditional probability
$p(\mathbf{y}~|~\mathbf{x})$ using SAR model but assumes the
labels to be independent of each other. The probability of each
label to be 1 (i.e., each pixel to be a target pixel) is
computed empirically from training data.
  \item  i-Auto model: considers the intensities to be iid Gaussian and models the the prior
$p(\mathbf{x})$ using Auto-Logistic MRF. The parameters for
Gaussian model for intensity are computed empirically from
training data.
\end{enumerate}

\begin{figure}[h]
\begin{center}
\subfigure[SAR-i model]{\includegraphics[width=1.5in,height=1.25in]{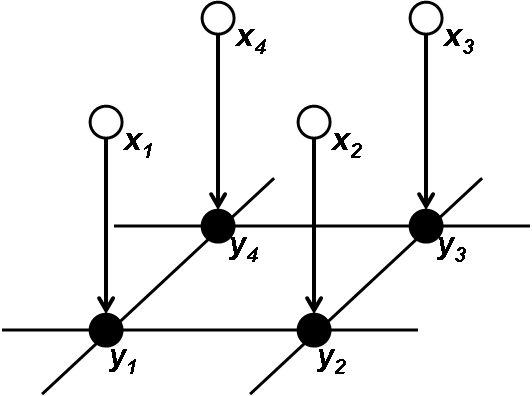}
}\qquad
\subfigure[i-Auto model]{\includegraphics[width=1.5in,height=1.25in]{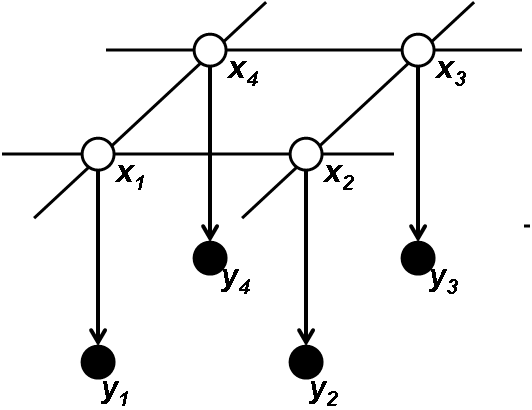}}
\end{center}
\caption{\footnotesize Bayesian networks for SAR-i and i-Auto.} \label{F:BN_RELAXED}
\end{figure}

That is, SAR-i and i-Auto assume independence in one of the two
layers that the proposed method considers to be dependent. The
Bayesian networks for these two models are illustrated in
Figure~\ref{F:BN_RELAXED}.

\begin{figure*}[t]
\begin{center}
\subfigure[IRD
01]{\includegraphics[width=2.2in,height=1.95in]{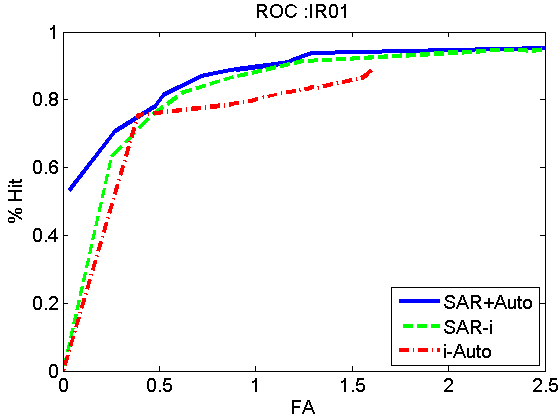}}
\subfigure[IRD
03]{\includegraphics[width=2.2in,height=1.95in]{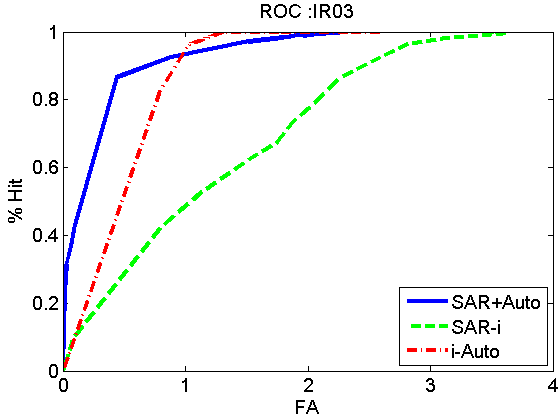}}
\subfigure[IRD
04]{\includegraphics[width=2.2in,height=1.95in]{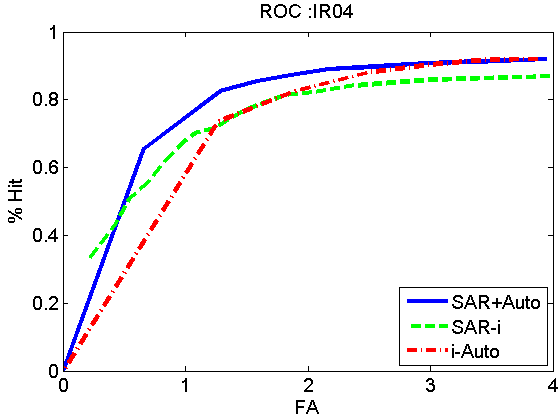}}
\subfigure[IRD
05]{\includegraphics[width=2.2in,height=1.95in]{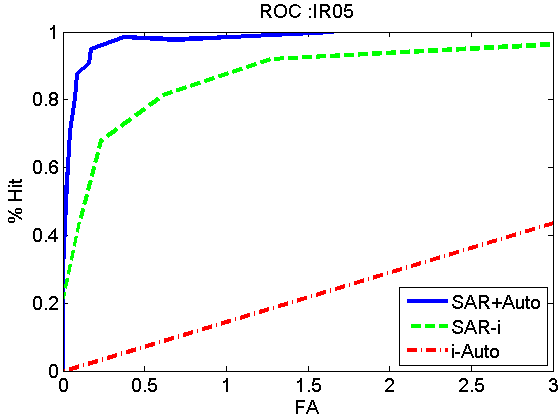}}
\end{center}
\caption{\footnotesize ROC curves of proposed (SAR-Auto, in blue
solid) and SAR-i (red dotted),
i-Auto (green dashed) methods. }
\label{F:RESULT2}
\end{figure*}

The performance comparison of these methods are shown in
Figure~\ref{F:RESULT2} where SAR-Auto refers to the proposed method.
The proposed method clearly exhibit a better performances than both
SAR-i and i-Auto methods for all datasets.

It is evident from the ROC curves that the proposed method SAR-Auto
achieves a high rate of detection, especially with FA rate $\le 1$
per frame. However, in IRD03 and IRD04, we observe occurrence of
many patches that has a high resemblance in intensity pattern with
target patches. That is why the proposed algorithm generates higher
rate of false alarms for IRD03 and IRD04. It is interesting to
observe that targets in these two image collections are
significantly brighter than the surrounding background. As a result,
the i-Auto model exhibits similar HIT and FA values of the coupled
SAR-Auto, i.e., pixelwise intensities independently are capable of
producing good detection rates.

\begin{figure}
\begin{center}
\subfigure[IRD01,
SAR-i]{\includegraphics[width=1.7in,height=1.4in]{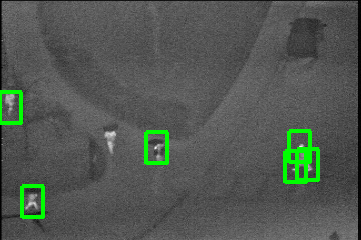}}
\subfigure[IRD01,
i-Auto]{\includegraphics[width=1.7in,height=1.4in]{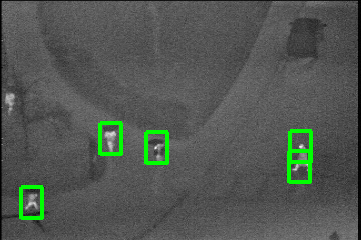}}
\subfigure[IRD01,
SAR-Auto]{\includegraphics[width=1.7in,height=1.4in]{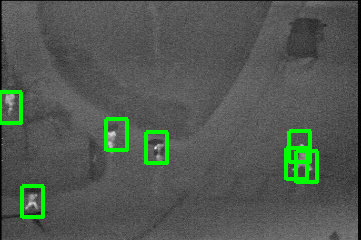}}
\subfigure[IRD04,
SAR-i]{\includegraphics[width=1.7in,height=1.4in]{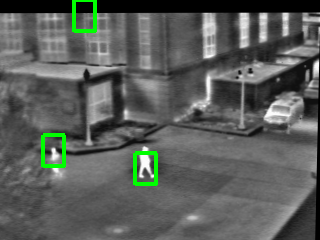}}
\subfigure[IRD04,
i-Auto]{\includegraphics[width=1.7in,height=1.4in]{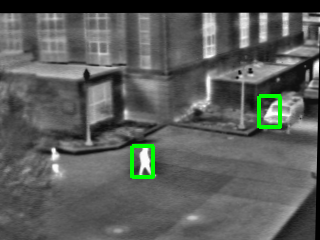}}
\subfigure[IRD04,
SAR-Auto]{\includegraphics[width=1.7in,height=1.4in]{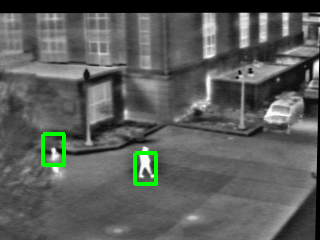}}
\subfigure[IRD05,
SAR-i]{\includegraphics[width=1.7in,height=1.4in]{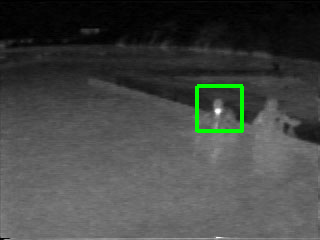}}
\subfigure[IRD05,
i-Auto]{\includegraphics[width=1.7in,height=1.4in]{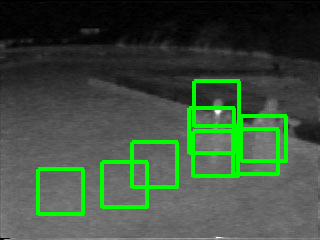}}
\subfigure[IRD05,
SAR-Auto]{\includegraphics[width=1.7in,height=1.4in]{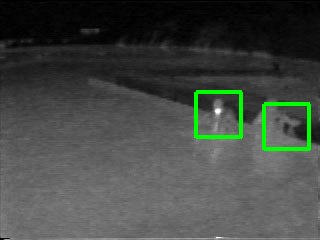}}
\subfigure[IRD03,
SAR-i]{\includegraphics[width=1.7in,height=1.4in]{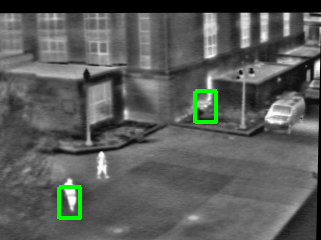}}
\subfigure[IRD03,
i-Auto]{\includegraphics[width=1.7in,height=1.4in]{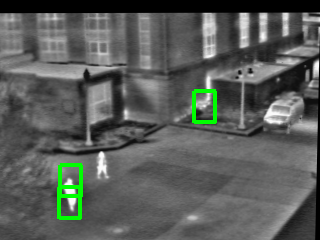}}
\subfigure[IRD03,
SAR-Auto]{\includegraphics[width=1.7in,height=1.4in]{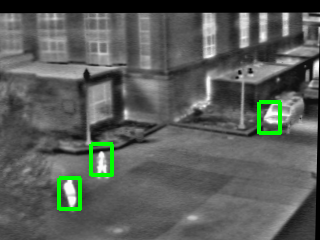}}
\end{center}
\caption{\footnotesize Sample input images for IR target
detection.}
\label{F:RESULT_QUAL}
\end{figure}

In datasets IRD01 and IRD05, the pixel intensities at individual
locations do not vary much between target and background classes.
However, the pattern of intensity arrangement within an image patch
is substantially different in these two classes. The i-Auto
approach, which models the intensities independently, generates many
incorrect detections in both IRD 01 and 05. On the other hand, the
SAR-i approach, which models the intensity correlation within the
image patch, performs better than i-Auto on these datasets. These
results suggest one needs to model both the label and intensity
dependence pattern in order to be robustly detect targets in varying
IR scenes. Sample qualitative results are compared in
Figure~\ref{F:RESULT_QUAL}. These outputs were generated by
operating at a false positive rate approximately 1 per frame for all
three techniques.

\subsection{Learned Parameters}
We display the parameter values learned for SAR-Auto model -- i.e.,
for SAR models of both target and background intensities and the
Auto-Logistic model of labels --  learned from  IRD05 dataset. The
left and right columns of Figure~\ref{E:SAR_CLIQ} displays the
parameters of the SAR models for target and background respectively.
The linear coefficients $\beta_{ij}^l$ for each clique is shown
above the dotted line in matrix form  and the mean and variances
$\mu^l, (\sigma^{l})^2$ are shown below. The neighborhood
relationship for background pixels (right column) are almost
symmetric in horizontal and vertical directions indicating similar
relationship between two neighbors in both both x and y-axis. On the
other hand, $\beta_{ij}^l$ for target is more informative as they
are asymmetric in both directions. The parameters $\nu_{i}$ and
$\gamma_{ij}$ learned for this dataset are $9.54$ and $-4.6924$
respectively (recall $\gamma_{ij}$ values are constrained to be the
same).

\begin{figure}
\begin{center}
\begin{equation*}
   \begin{array}{cc}
   \left [
       \begin{array}{ccc}
           & 0.044& \\
    0.443 &        &    0.479\\
           &  0.068 &
        \end{array}
   \right ] & \qquad
   \left [
       \begin{array}{ccc}
           & 0.016 & \\
     0.487&        & 0.483\\
           &  0.016 &        \\
        \end{array}
   \right ]  \\
   \dots \dots \dots \dots & \dots \dots \dots \dots\\
   117.4,~ 2.11 ~&~ 86.53,~ 1.19
   \end{array}
\end{equation*}
\end{center}
\caption{\footnotesize Learned parameters of SAR and Auto models.}
\label{E:SAR_CLIQ}
\end{figure}

\subsection{Incorporating background subtraction output}

Adding temporal information through background subtraction was
applied to datasets IRD 03 and IRD 04 both of which contain
temporally contiguous frames. As stated before, the output of the
MRF model outputs are combined with that of the background
subtraction algorithm by a logical AND operation. We found  this
simple operation performs well and reduces most of the false
positives. This is because (as is also elaborated at the end of this
section with an example), the false positives of the MRF model and
the background subtraction are almost always non-overlapping. We
used the implementation of~\cite{elgammal00} provided by the author
and selected the parameters of both the MRF and background
subtraction models to minimize the false negatives as much as
possible.

\begin{figure}[h]
\begin{center}
\subfigure[Input]{\includegraphics[width=1.75in,height=1.4in]{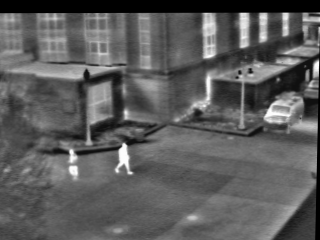}}\quad
\subfigure[SAR-Auto
output]{\includegraphics[width=1.75in,height=1.4in]{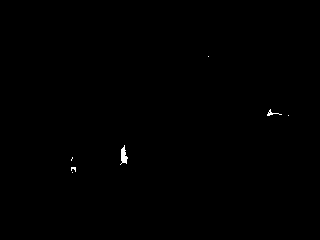}}\label{F:RESULT_BG_MRF}\\
\subfigure[BGsub
output]{\includegraphics[width=1.75in,height=1.4in]{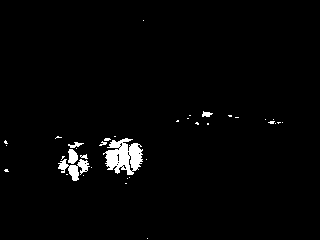}}\label{F:RESULT_BG_MRF}\quad
\subfigure[Combined
output]{\includegraphics[width=1.75in,height=1.4in]{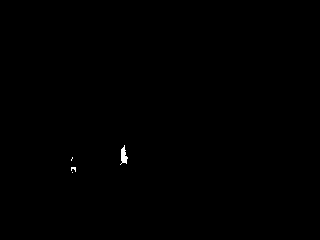}}\label{F:RESULT_BG_MRF}
\end{center}
\caption{\footnotesize Combining background subtraction with
SAR-Auto.}
\label{F:RESULT_BG}
\end{figure}

The output of each of the i-Auto, SAR-i and SAR-Auto
(Figure~\ref{F:RESULT_BG}(b)) approaches are combined with
background subtraction result (Figure~\ref{F:RESULT_BG}(c)) with a
pixelwise AND operator to generate the final output as shown in
Figure~\ref{F:RESULT_BG}(d). The numerical results of this process
is reported in Table~\ref{T:RESULT_BG}. For both the datasets, the
table shows, from left to right, the average percentage of true
positive and number of false positive per frame of~\cite{elgammal00}
alone (with parameter tuned to achieve 0 false negative rate), those
of each of the three aforementioned methods after adding background
subtraction (HIT-a, FA-a) and the average FA rate (FA-b)
corresponding to same detection accuracy without background
subtraction. Incorporation of background subtraction information
enables all three techniques to attain the largest possible true
positive rate with zero false alarm.

\begin{table}[h]
\scriptsize
\begin{tabular}{|c|c c|c c|c c|c c| } \hline
Dataset & \multicolumn{2}{c} {BGsub~\cite{elgammal00}} &
\multicolumn{2}{|c|} {SAR-Auto} & \multicolumn{2}{|c|} {i-Auto}
& \multicolumn{2}{|c|} {SAR-i} \\ \hline
& HIT& FA & HIT-a& FA-b $\rightarrow$ FA-a & HIT-a& FA-b $\rightarrow$ FA-a&  HIT-a&
FA-b $\rightarrow$ FA-a  \\ \hline
IRD03&100 & 15.57& 100 & 3.3 $\rightarrow$ 0 & 100 & 1.5 $\rightarrow$ 0 & 99.73 & 3.6467 $\rightarrow$ 0 \\ \hline
IRD04& 99.77&11.4 & 92.12& 3.94 $\rightarrow$ 0&92.1& 3.87 $\rightarrow$ 0&82.12& 8.55 $\rightarrow$ 0 \\
\hline
\end{tabular}
\caption{Performance improvement with background subtraction.}
\label{T:RESULT_BG}
\end{table}

\begin{figure}[h]
\begin{center}
\subfigure{\includegraphics[width=1.75in,height=1.4in]{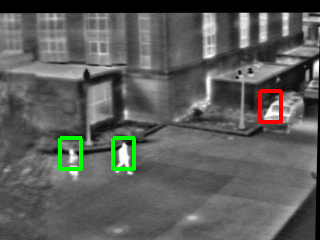}}
\subfigure{\includegraphics[width=1.75in,height=1.4in]{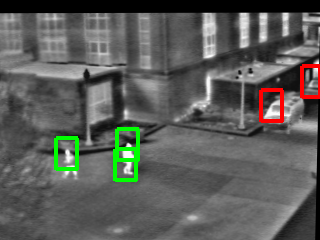}}
\subfigure{\includegraphics[width=1.75in,height=1.4in]{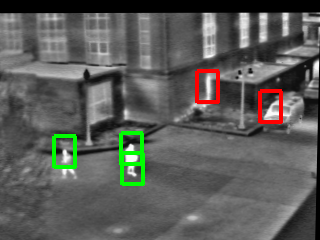}}
\end{center}
\caption{\footnotesize Sample false positives (red boxes) of
SAR-Auto.}
\label{F:RESULT_FA}
\end{figure}

There is an intuitive explanation why a logical AND operation
between combination of the proposed method (e.g., SAR-Auto) and
background subtraction results would remove the false positives. The
SAR-Auto method generates false positives in the background regions
that resembles the target pattern, and the intensities within does
not change significantly i.e., typically static regions, for a
period of time (shown in red in Figure~\ref{F:RESULT_FA}).
Background subtraction methods, by construction, would identify
these regions as background. Therefore, the false detections from
MRF models and the background subtraction method generally do not
overlap with each other.

\section{Conclusion}\label{S:CONC}

This study demonstrates the advantage of enforcing neighborhood
dependence in both intensity and label layers for IR target
detection. The intensity and label value at any pixel is related to
those of its neighbors by SAR and Auto-Logistic MRF models
respectively. High accuracies achieved in our experiments suggest
these models were able to learn the local image patterns adequately
enough for locating object regions in IR image. We show that
independence assumption in any of the two layers (intensity and
label) would deteriorate the performance, i.e., coupling two MRF
models with each other is necessary for satisfactory performance.
Furthermore, we combine temporal information into the framework to
improve the detection accuracy. We strongly believe findings in this
paper will provide new insight and instigate research in utilizing
patch appearance and the dependence among the pixels within for
target detection in infra-red imagery.

\bibliographystyle{plain}
\bibliography{cviu12mrf}

\end{document}